\documentclass[letterpaper]{article} 
\usepackage{aaai2026}  
\usepackage{times}  
\usepackage{helvet}  
\usepackage{courier}  
\usepackage[hyphens]{url}  
\usepackage{graphicx} 
\usepackage{natbib}  
\usepackage{caption} 
\usepackage{algorithm}
\usepackage{algorithmic}
\usepackage{amsmath}
\usepackage{tabularx}
\usepackage{multirow}
\usepackage{newfloat}
\usepackage{listings}
\DeclareCaptionStyle{ruled}{labelfont=normalfont,labelsep=colon,strut=off} 
\floatstyle{ruled}
\newfloat{listing}{tb}{lst}{}
\floatname{listing}{Listing}
\title{FGNet: Leveraging Feature-Guided Attention to Refine SAM2 for 3D EM Neuron Segmentation}
\author{
    Zhenghua Li\textsuperscript{\rm 1, \rm 2},
    Hang Chen\textsuperscript{\rm 1, \rm 2},
    Zihao Sun\textsuperscript{\rm 3},
    Kai Li\textsuperscript{\rm 1, \rm 2},
    Xiaolin Hu\textsuperscript{\rm 1, \rm 2, \rm 4}\thanks{Corresponding author.}
}
\affiliations{
    
    \textsuperscript{\rm 1}Department of Computer Science and Technology, Institute for AI, BNRist, Tsinghua University, Beijing 100084, China\\
    \textsuperscript{\rm 2}Tsinghua Laboratory of Brain and Intelligence (THBI), IDG/McGovern Institute for Brain Research, Tsinghua University, Beijing 100084, China\\
    \textsuperscript{\rm 3}Zhili College, Tsinghua University, Beijing 100084, China\\
    \textsuperscript{\rm 4}Chinese Institute for Brain Research (CIBR), Beijing 100010, China\\
    \{li-zh24, chenhang20, szh24, li-k24\}@mails.tsinghua.edu.cn, xlhu@tsinghua.edu.cn
    
}

\usepackage{bibentry}

\begin{document}

\maketitle

\begin{abstract}
Accurate segmentation of neural structures in Electron Microscopy (EM) images is paramount for neuroscience. However, this task is challenged by intricate morphologies, low signal-to-noise ratios, and scarce annotations, limiting the accuracy and generalization of existing methods. To address these challenges, we seek to leverage the priors learned by visual foundation models on a vast amount of natural images to better tackle this task. Specifically, we propose a novel framework that can effectively transfer knowledge from Segment Anything 2 (SAM2), which is pre-trained on natural images, to the EM domain. We first use SAM2 to extract powerful, general-purpose features. To bridge the domain gap, we introduce a Feature-Guided Attention module that leverages semantic cues from SAM2 to guide a lightweight encoder, the Fine-Grained Encoder (FGE), in focusing on these challenging regions. Finally, a dual-affinity decoder generates both coarse and refined affinity maps. Experimental results demonstrate that our method achieves performance comparable to state-of-the-art (SOTA) approaches with the SAM2 weights frozen. Upon further fine-tuning on EM data, our method significantly outperforms existing SOTA methods. This study validates that transferring representations pre-trained on natural images, when combined with targeted domain-adaptive guidance, can effectively address the specific challenges in neuron segmentation.
\end{abstract}

\section{Introduction}
\label{intro}

Neuron segmentation in electron microscopy (EM) images is important to neuroscience research, as accurate segmentation of complex neural structures is crucial for understanding brain connectivity and function~\cite{cremi,ac3/ac4,lsd}. However, this task remains highly challenging due to the inherent characteristics of EM data, 3D EM neuron images typically exhibit intricate morphological structures, low signal-to-noise ratios, and massive neuron populations, making it difficult to capture fine-grained details and maintain segmentation consistency~\cite{lee_superhuman_2017}.

Existing approaches to EM neuron segmentation can be roughly divided into two categories: (1) \textit{Methods without pre-training}, which are trained from scratch on datasets specifically tailored for EM segmentation, including PEA~\cite{pea}, LSD~\cite{lsd}, and CAD~\cite{cad}. (2) \textit{Methods with EM-specific pre-training}, which aim to enhance generalization through pre-training strategies, including typically self-supervised learning such as DbMIM~\cite{dbmim} and EMmamba~\cite{tokenunify}.
However, all the aforementioned models are trained exclusively on EM data, which inherently limits their performance due to the scarcity of labeled EM datasets, as annotating complex 3D neural structures is extraordinarily difficult and labor-intensive.

Meanwhile, with the rise of pre-trained foundation models, significant progress has been made in the field of natural image segmentation. Models such as Mask2Former~\cite{mask2former}, SAM~\cite{sam}, and SAM2~\cite{sam2},  trained on large-scale natural image datasets with abundant segmentation masks, have demonstrated exceptional feature extraction capabilities and strong generalization abilities. \textit{This raises a question: can the powerful representations learned from natural images be transferred to EM neuron segmentation to alleviate the data scarcity issue in the EM domain? }

In this paper, we try to answer this question. Considering that SAM2 is currently the latest and most widely used segmentation model, pre-trained on a large number of datasets, we mainly conducted our experiments on it. Previous work in medical image segmentation has utilized foundation models by fine-tuning or injecting adapters~\cite{sam-adapter}. However, with these techniques we found that the results were not satisfactory (see Section~\ref{sec:cmp adaption}). 
This challenge stems from the highly complex nature of neuronal image details. Architectures like SAM2, which incorporate multiple downsampling layers in their design, may inadvertently compromise some fine-grained structural information during image processing.
\textit{This observation highlighted the need for a dedicated refinement stage to recover lost details.}

To better incorporate fine-grained information by extracting details from the original image, existing works such as SAM-REF~\cite{sam-ref} adopt a strategy combining global and local refinement, targeting regions with high error rates, while BPR~\cite{look-ref} employs boundary patch refinement, focusing specifically on boundary regions. Yet these strategies have a critical limitation in EM neuron segmentation: due to the large number of neuron instances in EM images, it is difficult to determine which specific regions need refinement in advance, making such methods unsuitable for this task. \textit{This limitation motivates us to explore the possibility of refining the results directly using a separate, specialized network.}

Consequently, we design a lightweight network for refinement. To ensure effective results with minimal parameters, we develop a feature-guided attention module. This module guides the refinement network by leveraging features from the foundation model. Its purpose is to direct the network to extract fine-grained information from the original image, specifically the details that the foundation model may have overlooked during its downsampling stages.

Extensive experiments demonstrated that our methods achieved competitive performance even when SAM2's weights were frozen. Furthermore, with moderate fine-tuning on EM data, the proposed method outperformed state-of-the-art methods, providing a practical solution for balancing transfer efficiency and segmentation accuracy.

\section{Related Work}

\subsection{Electron Microscopy Neuron Segmentation}

Most methods involve designing expert models trained from scratch on specific datasets. The dominant approach predicts pixel affinities (encoding connectivity), followed by the watershed algorithm for fragment generation and aggregation to form complete neurons. Related improvements include PEA \cite{pea} using contrastive learning to enhance the robustness of affinities, and CAD \cite{cad} distilling 3D knowledge into 2D networks for efficiency. FragViT \cite{fragvit} utilizes a vision Transformer with hierarchical aggregation, and APViT \cite{apvit} has a refinement module with appearance prompts. Other expert models within this paradigm include FFN \cite{ffn}, which segments by iteratively growing neuron fragments but is two orders of magnitude slower than affinity-based methods \cite{agq}; AGQ designs affinity-guided query initialization combined with learnable queries to achieve end-to-end segmentation but requires adjusting the number of queries for specific data.

Beyond expert models, EM-specific pre-training strategies have emerged, such as self-supervised pre-training in DbMIM \cite{dbmim} and long-range modeling in EMmamba \cite{tokenunify}, which can improve generalization ability but suffer from limited EM data.

\subsection{Adaptation from Foundation Models}
Foundation models like Segment Anything Model (SAM) \cite{sam} and its successor SAM2 \cite{sam2} have demonstrated remarkable capabilities, trained on large-scale datasets and exhibiting strong zero-shot performance across diverse segmentation tasks \cite{roy2023sam}. However, despite their strong zero-shot capabilities, they still underperform on specific downstream tasks.
MedSAM \cite{medsam} adapts SAM to medical segmentation by freezing the prompt encoder and fine-tuning the image encoder/mask decoder for domain alignment. Medical SAM Adapter \cite{medsamadapter} uses lightweight adapters for parameter-efficient adaptation without full fine-tuning.
SemiSAM \cite{semisam} leverages SAM as an auxiliary supervision branch, generating pseudo-labels for semi-supervised learning. SAM-Med3D \cite{medsam3d} extends SAM to 3D by converting core components for volumetric medical data. SAM-REF \cite{sam-ref} improves quality via global/local refinement.
These methods demonstrate various strategies for adapting and refining foundation models for specialized tasks.

\subsection{Attention Modules in Segmentation and Detection}
Attention mechanisms enhance feature representation by focusing on critical regions, with diverse modules tailored for segmentation and detection: Woo et al. \cite{woo2018cbam} proposed CBAM, a lightweight module that infers channel and spatial attention maps sequentially; Hu et al. \cite{hu2018squeeze} introduced SENet, a pioneering channel attention framework modeling inter-channel dependencies via global pooling and fully connected layers; Wang et al. \cite{wang2018non} extended attention to global contexts with Non-local Networks, capturing long-range dependencies; Xu et al. \cite{xu2019ac} presented AC-FPN, which uses Context and Content Attention Modules to enhance multi-scale fusion; Fu et al. \cite{fu2019dual} designed DANet, combining position and channel attention to model global dependencies; Yang et al. \cite{yang2019dea} proposed DEA-Net with Content-Guided Attention; Wang et al. \cite{wang2020eca} developed ECA-Net, an efficient channel attention variant using 1D convolutions; Li et al. \cite{li2024vl} introduced VL-SAM, leveraging attention maps as prompts for open-vocabulary segmentation. These modules cover channel, spatial, and global attention, demonstrating versatility in feature refinement, and our work builds on this by using pre-trained weights to guide attention modules in extracting fine-grained details.

\begin{figure*}[t]
\centering
\includegraphics[width=\textwidth]{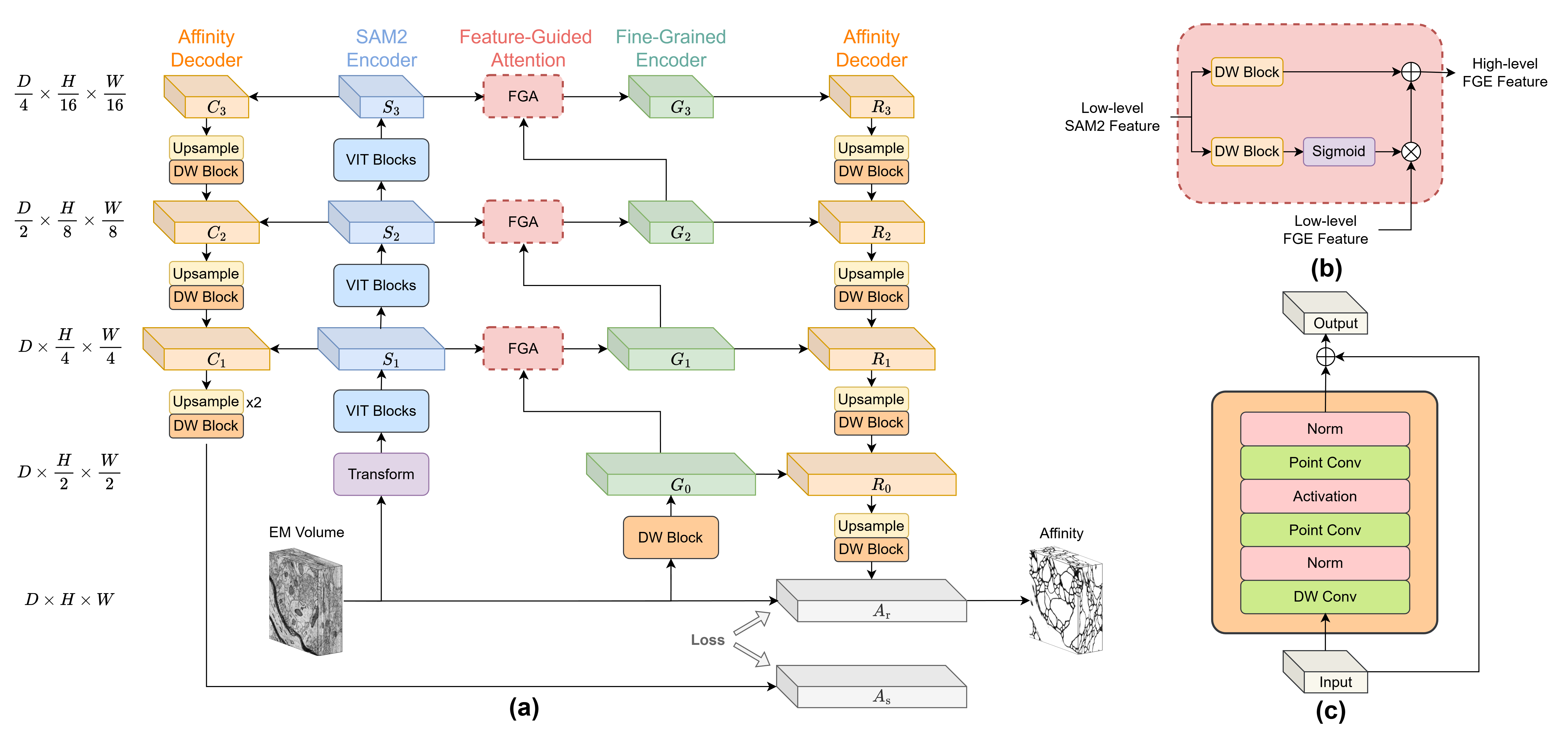}
\caption{Network architecture. (a) Four core components: SAM2 Encoder, Feature-Guided Attention (FGA), Fine-Grained Encoder (FGE), and a pair of Affinity Decoders. Modules of different types are represented with distinct colors. The left column indicates the size of each feature in the horizontal dimension. (b) Detailed structure of the FGA module. The plus sign ($\oplus$) represents element-wise addition, and the multiplication sign ($\otimes$) represents element-wise multiplication. (c) Detailed structure of the DW Block.}
\label{fig:network}
\end{figure*}
\section{Methods}
\label{sec:methods}

\subsection{Overall Pipeline}
\label{sec:overall pipeline}
Let the volume of electron microscopy (EM) neuron images be denoted as $V \in \mathcal{R}^{D \times H \times W}$, where $D$, $H$, and $W$ represent the depth, height, and width dimensions, respectively. Our segmentation model predicts affinity matrices via a hierarchical pipeline (Figure~\ref{fig:network}(a)) with four core components: SAM2 Encoder, Feature-Guided Attention (FGA), Fine-Grained Encoder (FGE), and dual affinity decoders (shared architecture, distinct inputs). The SAM2 Encoder processes \( V \) to generate multi-scale features \(S\), which are fed into the first decoder to produce a coarse affinity matrix \( A_\text{s} \). FGA then modulates \( S \) to guide FGE in generating fine-grained features \( G \), which the second decoder uses to derive a refined matrix \( A_\text{r} \). Both \( A_\text{s} \) and \( A_\text{r} \) are supervised against \( A_\text{gt} \). 

During inference, the affinity decoder on the left in Figure~\ref{fig:network}(a) is not used, while affinity \( A_\text{r} \) is then processed using watershed~\cite{mala} and agglomeration~\cite{mala} algorithms to generate the final segmentation result $O \in \mathcal{R}^{D \times H \times W}$.

\subsection{SAM2 Encoder}  
The input EM volume is first processed by a transformation module to align with SAM2's input specifications, followed by a sequence of ViT Blocks~\cite{sam2} to generate hierarchical features \( S_i \) (with \( i \in \{1,2,3\} \) indicating distinct levels). This component is visualized as the blue section in Figure~\ref{fig:network}(a).

\subsection{FGA Module and Fine-Grained Encoder}

The FGE extracts fine-grained features \( G_i \) (with \( i \in \{0,1,2,3\} \) indicating distinct levels) from the original EM volume \( V \), guided by the SAM2 Encoder’s hierarchical features \( S_i \) through the FGA module. Specifically, the initial fine-grained feature \( G_0 \) is derived directly from \( V \) as:  
\begin{equation}
G_0 = \text{DW}(V),
\end{equation}  
where \( \text{DW}(\cdot) \) denotes the depth-wise convolution block (abbreviated as DW Block), whose structure is depicted in Figure~\ref{fig:network}(c).  

The FGA module refines subsequent features \( G_i \) from the preceding \( G_{i-1} \) under \( S_i \) guidance, employing a two-branch attention mechanism (Figure~\ref{fig:network}(b)). For each level \( i \), SAM2 Feature \( S_i \) first generates two attention maps \( a_i \) and \( b_i \) to capture distinct structural priors:  
\begin{equation}
a_i = \sigma\left(\text{DW}(S_i)\right),
\label{eq:a}
\end{equation}  
\begin{equation}
b_i = \text{DW}(S_i),
\end{equation}  
where \( \sigma(\cdot) \) is the sigmoid activation function constraining \( a_i \in [0,1] \).  
The refined feature \( G_i \) is computed by
\begin{equation}
G_i = a_i \odot G_{i-1} + b_i,
\end{equation}  
where \( \odot \) denotes element-wise multiplication. This design enables \( a_i \) to weight informative regions in \( G_{i-1} \) and \( b_i \) to compensate for feature biases.

\subsection{Affinity Decoder}
We employ two affinity decoders with identical architecture but varying depths and inputs, as shown in Figure~\ref{fig:network}(a).

The left decoder processes the hierarchical features \( S_i \) from the SAM2 Encoder to predict the coarse affinity matrix \( A_\text{s} \). It consists of a sequence of upsampling operations and DW Blocks, which progressively restore the feature resolution to match the input volume dimensions.  

The right decoder, sharing the same structure design but with a different number of layers, takes the fine-grained features \( G_i \) as input. Through a parallel sequence of upsampling steps and DW Blocks, it generates the refined affinity matrix \( A_\text{r} \). Both decoders leverage DW Blocks to preserve computational efficiency while maintaining computational efficiency, with their distinct layer counts tailored to the characteristics of their respective input features (\( S_i \) for coarse prediction and \( G_i \) for fine-grained refinement).

Notably, the left Affinity Decoder functions as a deep supervision branch. It provides supervisory signals to enable SAM2's feature extraction capabilities to adapt as much as possible to EM data. However, its output remains coarse due to its primary focus on initial feature adaptation. In contrast, the right decoder produces refined results by extracting fine-grained features under the guidance of the FGA module, thus serving as the final affinity result.
\section{Experiments}
\label{sec:experiments}

\subsection{Datasets and Metrics}
\label{subsec:datasets_metrics}

We evaluated our method on three widely-used public electron microscopy (EM) neuron segmentation datasets:

\begin{itemize}
    \item \textbf{AC3/AC4} \cite{ac3/ac4}: Labeled subsets derived from the mouse somatosensory cortex dataset. The dataset images are acquired at a spatial resolution of $3\times3\times29$ $\text{nm}^3$. Specifically, the AC3 subset consists of 256 sequential sections, while the AC4 subset contains 100 sequential sections. For evaluation, the data partitioning follows the protocol in \cite{agq}: the first 80 sections of AC4 are allocated for training, the subsequent 20 sections of AC4 for validation, and the first 100 sections of AC3 for testing, with each section in both subsets having dimensions of $1024\times1024$ pixels.
    \item \textbf{CREMI} \cite{cremi}: A benchmark dataset containing 3D EM volumes of Drosophila melanogaster brain tissue, with a resolution of $4\times4\times40$ $\text{nm}^3$. The dataset includes 3 volumes (A, B, C) each with dimensions $1250\times1250\times125$, and provides manually annotated segmentation masks for training and evaluation. Every volume is divided into 100 sections for training and 25 sections for testing.
    \item \textbf{Wafer4} \cite{cad}: Collected from a region of the mouse medial entorhinal cortex, which has a size of 1250×1250×125 voxels. This dataset is divided into 100 sections for training and 25 sections for testing.
\end{itemize}
We utilized two widely adopted metrics to quantitatively evaluate segmentation results, both of which measure error (where lower values indicate better performance): 
\begin{itemize}  
    \item \textbf{Variation of Information (VOI)} : A dissimilarity metric that quantifies the discrepancy between predicted and ground truth segmentations\cite{voi}, explicitly accounting for both over-segmentation and over-merging errors. It decomposes into two components: \( \text{VOI}_{\text{split}} \) (assessing over-segmentation) and \( \text{VOI}_{\text{merge}} \) (assessing over-merging).
    \item \textbf{Adapted Rand Error (Arand)} : An adjusted variant of the Rand Index\cite{arand}, optimized to address the uneven distribution of object sizes in EM image segmentation tasks. This metric measures the disagreement between predicted results and ground truth.  
\end{itemize}
\subsection{Implementation Details}
\label{subsec:implementation_details}
The input block size for both training and inference was set to \(16 \times 256 \times 256\), with a stride of \(8 \times 128 \times 128\) used for inference. The model was optimized via the Adam optimizer~\cite{kingma2014adam} with an initial learning rate of \(1\times 10^{-4}\), training was conducted with a batch size of 4 over 10,000 iterations. Our framework was implemented using \text{pytorch\_connectomics} codebase~\cite{pytorch_connectomics}. Our experiments were conducted trained on 4 NVIDIA 3090 GPUs, each with 24GB memory.

\begin{table*}[t]
\caption{Comparison of different models on AC3/AC4 datasets. Lower values are better for all metrics. $^*$ indicates results are reproduced by us. $\dagger$ denotes a modified version by the original authors. VOI results are obtained by the Waterz~\cite{mala} post-processing.}
\centering
\small
\begin{tabular}{c|l|l|ccc|c}
\hline
\multicolumn{2}{c|}{Model}                                          & Reference     & $\rm VOI_{split}$ & $\rm VOI_{merge}$ &  $\rm VOI$ & Arand \\ \hline \hline
\multirow{13}{*}{No Pre-training} & ResUNet~\cite{3d_unet}                        & MICCAI'16   & 1.037$^*$      & 0.258$^*$      & 1.295$^*$    & 0.154$^*$ \\
&SuperHuman~\cite{lee_superhuman_2017}         & ArXiv'17    & 1.145$^*$      & 0.263$^*$     & 1.408$^*$    & 0.122$^*$ \\
&MALA~\cite{mala}                              & TPAMI'19    & 1.304$^*$      & 0.242$^*$      & 1.546$^*$    & 0.120$^*$ \\
&SEUNet~\cite{pytorch_connectomics}            & ArXiv'21    & 1.031$^*$      & 0.251$^*$      & 1.282$^*$    & 0.156$^*$ \\
& SwinUNETR~\cite{swinunetr}   &MICCAI'21   & 1.238$^*$      & 0.191$^*$      & 1.429$^*$    & 0.110$^*$  \\
&PEA~\cite{pea}                                & AAAI'22     & 0.852$^*$      & 0.232$^*$      & 1.084$^*$    & 0.094$^*$ \\ 

&UNETR~\cite{unetr}                    & WACV'22     & 1.048$^*$      & 0.237$^*$      & 1.285$^*$    & 0.116$^*$  \\
&LSD~\cite{lsd}                                & NM'23       & 1.448$^*$      & 0.229$^*$      & 1.677$^*$    & 0.134$^*$ \\
&APViT~\cite{apvit}                    & IJCAI'23    & 0.767      & 0.204      & 0.971    & 0.078 \\
&FragViT~\cite{fragvit}                & AAAI'24     & 0.868      & 0.191      & 1.054    & 0.093 \\

&CAD~\cite{cad}                                & CVPR'24     & 0.601	    & 0.431	    & 1.032	    & 0.119 \\ 
&CAD + KD~\cite{cad}                           & CVPR'24     & 0.533             & 0.351             & 0.884     & 0.081 \\
&AGQ~\cite{agq}             & ICLR'25    &  0.677    &   0.290    &    0.967  & 0.095 \\ \hline
\multirow{3}{*}{Pre-trained on EM} 
&DbMIM+UNETR~\cite{dbmim}   & IJCAI'23    & 0.647      & 0.285      & 0.931   & 0.243 \\
&SegNeuron~\cite{segneuron} & MICCAI'24     & 0.698$^*$      & 0.245$^*$    &0.943$^*$     &  0.088$^*$  \\
&EMmamba~\cite{tokenunify}             & ICCV'25    & 0.938      & 0.863      & 1.801    & 0.284 \\
&EMmamba$^\dagger$~\cite{tokenunify}   & ICCV'25    & 0.448      & 0.544      & 0.992    & 0.137 \\
      
\hline
\multirow{2}{*}{Pre-trained on Natural Images}
&FGNet w/o finetune SAM2                                           &  -            &  0.647     &   0.263    & 0.910   & 0.096 \\ 
&FGNet                                          &  -            & 0.614      &  0.183     & \textbf{0.797}    & \textbf{0.069} \\ \hline
\end{tabular}
\label{tab:ac3_ac4}
\end{table*}

\begin{table*}[ht]
\small
\setlength{\tabcolsep}{1.3mm}
\centering
\caption{Results on the CREMI dataset. Results on three volumes are reported. Note that VOI is the overall metric of $\rm VOI_{s}$ and $\rm VOI_{m}$. Results were borrowed from ~\cite{cad}. The best and second-best results are bolded and underlined, respectively.}
\begin{tabular}{l|cccc|cccc|cccc}
\hline
                 & \multicolumn{4}{c|}{CREMI-A}                       & \multicolumn{4}{c|}{CREMI-B}                       & \multicolumn{4}{c}{CREMI-C}                        \\ \hline \hline
Model            & $\rm VOI_{s}$ & $\rm VOI_{m}$  & \multicolumn{1}{c|}{VOI}   & Arand & $\rm VOI_{s}$ & $\rm VOI_{m}$  & \multicolumn{1}{c|}{VOI}   & Arand & $\rm VOI_{s}$ & $\rm VOI_{m}$  & \multicolumn{1}{c|}{VOI}   & Arand \\ \hline
SuperHuman~\cite{lee_superhuman_2017}       & 0.399 & 0.241 & \multicolumn{1}{c|}{0.640} & 0.089 & 0.554 & 0.222 & \multicolumn{1}{c|}{0.776} & 0.048 & 0.820 & 0.338 & \multicolumn{1}{c|}{1.158} & 0.179 \\
MALA~\cite{mala}             & 0.398 & 0.236 & \multicolumn{1}{c|}{0.634} & 0.085 & 0.589 & 0.261 & \multicolumn{1}{c|}{0.850} & 0.041 & 0.842 & 0.332 & \multicolumn{1}{c|}{1.174} & 0.162 \\
PEA~\cite{pea}              & 0.329 & 0.298 & \multicolumn{1}{c|}{0.626} & 0.091 & 0.411 & 0.374 & \multicolumn{1}{c|}{0.785} & 0.041 & 0.745 & 0.446 & \multicolumn{1}{c|}{1.191} & 0.169 \\
APViT~\cite{apvit}            & 0.445 & 0.260 & \multicolumn{1}{c|}{0.704} & 0.117 & 0.579 & 0.201 & \multicolumn{1}{c|}{0.781} & \underline{0.032} & 0.884 & 0.234 & \multicolumn{1}{c|}{1.118} & 0.110 \\
          
DbMIM+UNETR~\cite{dbmim}            & 0.411 &  0.331 & \multicolumn{1}{c|}{0.743} & 0.131 & 0.642 & 0.381 & \multicolumn{1}{c|}{1.023} & 0.092& 0.925 & 0.276& \multicolumn{1}{c|}{1.201} & \underline{0.107} \\
CAD~\cite{cad}              & 0.326 & 0.299 & \multicolumn{1}{c|}{0.625} & 0.107 & 0.402 & 0.347 & \multicolumn{1}{c|}{0.749} & 0.045 & 0.738 & 0.455 & \multicolumn{1}{c|}{1.193} & 0.170 \\
CAD + KD~\cite{cad}         & 0.313 & 0.252 & \multicolumn{1}{c|}{\underline{0.565}} & \underline{0.079} & 0.379 & 0.305 & \multicolumn{1}{c|}{\underline{0.684}} & \textbf{0.030} & 0.738 & 0.322 & \multicolumn{1}{c|}{\underline{1.060}} & 0.149 \\

\hline
FGNet w/o finetune SAM2&  0.393&0.250  &  \multicolumn{1}{c|}{0.643} & 0.101  & 0.398 & 0.380 & \multicolumn{1}{c|}{0.778}  & 0.061 & 0.901 &  0.299 & \multicolumn{1}{c|}{1.201} & 0.149 \\ 
FGNet  & 0.331 & 0.230 & \multicolumn{1}{c|}{\textbf{0.561}} & \textbf{0.075}  &0.514  & 0.163 & \multicolumn{1}{c|}{\textbf{0.677}}  & 0.040 & 0.540 &  0.481 & \multicolumn{1}{c|}{\textbf{1.021}} & \textbf{0.081} \\ \hline
\end{tabular}
\label{tab:cremi}
\end{table*}

\begin{table}[ht]
\centering
\small
\setlength{\tabcolsep}{2mm}
\caption{Results on the Wafer4 dataset.
Results were obtained from ~\cite{cad}.
The best results are bolded.
}
\begin{tabular}{l|ccc|c}
\hline
Model               & $\rm VOI_{s}$ & $\rm VOI_{m}$  & VOI   & Arand \\ \hline
SuperHuman~\cite{lee_superhuman_2017}  & 0.452      & 0.166      & 0.618    & 0.041         \\
MALA~\cite{mala}       & 0.455      & 0.158      & 0.613    & 0.036         \\
PEA~\cite{pea}        & 0.421      & 0.172      & 0.593    & 0.034         \\
APViT~\cite{apvit}      & 0.581      & 0.123      & 0.704    & 0.036         \\
CAD~\cite{cad}                 & 0.404      & 0.224      & 0.627    & 0.051         \\
CAD + KD~\cite{cad}            & 0.415      & 0.144      & 0.559    & 0.030         \\ \hline
FGNet w/o finetune SAM2     &  0.598     &  0.131     & 0.729    & 0.042         \\ 
FGNet    &  0.412     &  0.121     & \textbf{0.533}    & \textbf{0.028}         \\ \hline
\end{tabular}
\label{tab:wafer4}
\end{table}


\subsection{Quantitative Results}

To thoroughly evaluate the effectiveness of the proposed method, we compared our method with state-of-the-art EM neuron segmentation methods on the AC3/AC4, CREMI and Wafer4 datasets, respectively. Our approach exhibited notable performance characteristics across different training configurations:  

Our model achieved state-of-the-art (SOTA) performance across multiple datasets (see Tables \ref{tab:ac3_ac4}, \ref{tab:cremi}, and \ref{tab:wafer4}).
Notably, on the AC3/AC4 datasets, our method demonstrated strong performance in the VOI metric compared to the previous SOTA method CAD \cite{cad}. Notably, our model achieved performance comparable to CAD, even without fine-tuning the pre-trained SAM2 encoder. Furthermore, after fine-tuning, our approach achieved a substantial 12.5\% improvement over CAD, solidifying its superior performance. This superior performance extended to other benchmarks: on CREMI A, B, and C datasets, our fine-tuned model outperformed existing SOTA methods by 0.7\%, 1.0\%, and 3.7\% respectively in VOI score; on the Wafer4 dataset, we observed a notable 4.6\% improvement in VOI score.

These results validated the effectiveness of our framework in balancing large-scale pre-training advantages with dataset-specific optimization, enabling consistent performance boosts across diverse EM imaging benchmarks.

\subsection{Qualitative Results}
\label{ssec:visualization_results}

\begin{figure*}[tb]
\centering
\includegraphics[width=\textwidth]{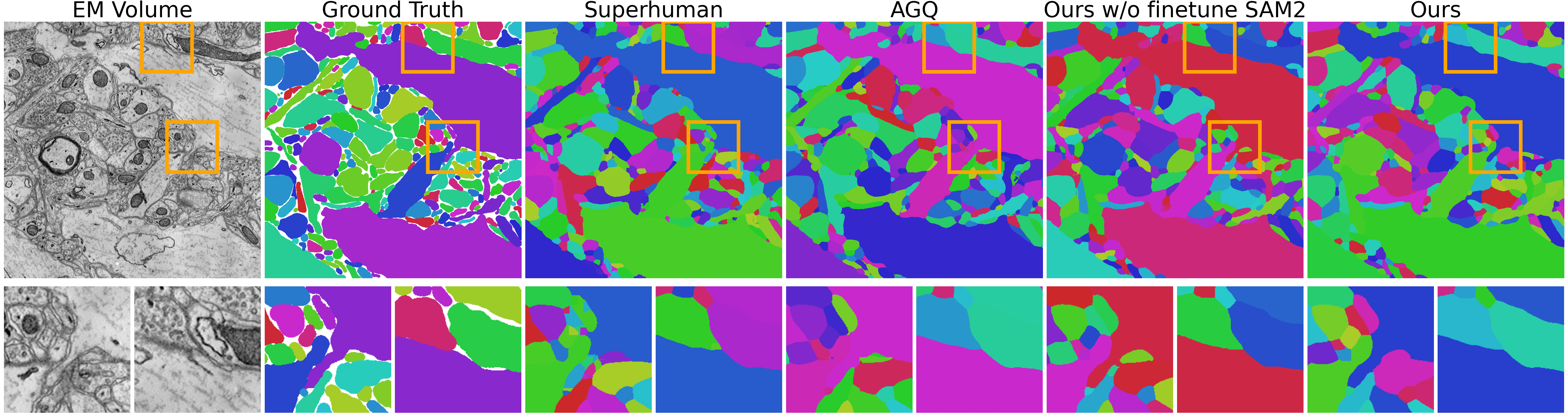}
\caption{
2D visualization of segmentation results obtained from different methods.
For each result, the first row presents a 2D slice and the second row shows two zoomed-in regions.
Best viewed in digital with zoom-in.
}
\label{fig:show2d}
\end{figure*}

\begin{figure*}[tb]
\centering
\includegraphics[width=0.94\textwidth]{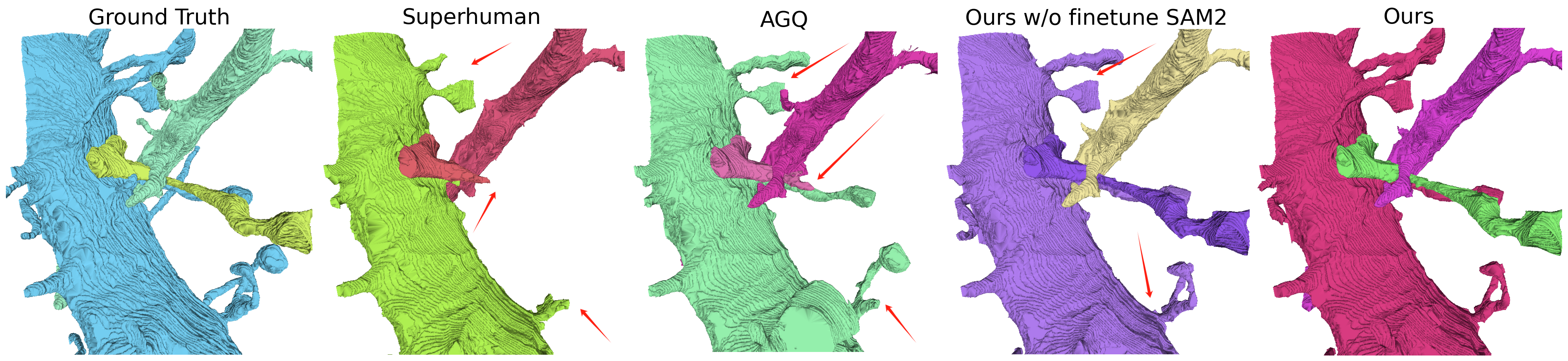}
\caption{
3D visualization of segmentation results obtained from different methods.
The 3D neuron morphology is shown, with red arrows indicating some of the slender dendrites in the neuron.
}
\label{fig:show3d}
\end{figure*}

We visualize the results of the proposed method in comparison with those of previous methods, showcasing both the 2D sections (In Figure \ref{fig:show2d}) and the 3D neuron morphology (In Figure \ref{fig:show3d}). 

The 2D slices present pixel-level segmentation results. As highlighted by the orange boxes, previous methods exhibit numerous over-segmentation and under-segmentation errors, whereas our method successfully separates these regions. In the 3D morphology visualization, missegmented areas in previous methods are highlighted with red arrows, particularly where multiple fine-scale synapses of the main neuron were not accurately predicted. In contrast, our method achieves a more precise reconstruction of the neuron's structure.

\subsection{Comparison of Foundation Model Adaptation Methods}
\label{sec:cmp adaption}
As we mentioned in Section~\ref{intro}, we first compared different adaptation methods for foundation models. The following are the adaptation approaches for different foundation models such as SAM and SAM2, which we have implemented and applied to EM segmentation for comparison. Specific implementation details can be found in the Appendix. 
\textit{Frozen}: Direct transfer with most pre-trained parameters frozen \cite{roy2023sam};
\textit{Fine-tuning}: Full fine-tuning of the foundation model \cite{medsam};
\textit{Refinement}: Direct integration of a learnable lightweight refinement network \cite{sam-ref}.

Experimental results in Table~\ref{tab:way_ablation} demonstrate that our method, by incorporating a feature-guided attention mechanism between the foundation model and the lightweight network, outperforms all the above approaches.
This is primarily because the substantial domain gap between natural images and EM data, coupled with significant discrepancies in signal-to-noise ratio, renders both frozen and naively fine-tuned models ineffective at capturing fine-grained details. While adding an extra network yields better refinement, exclusive reliance on refinement without leveraging the general feature guidance from the foundation model overlooks its supervisory value in aligning with domain-agnostic patterns.
These results further confirmed that the integration of both fine-tuning and an adaptive lightweight network (as enabled by our design) is essential for achieving optimal performance in EM neuron segmentation.
\begin{table}[t]
\caption{Comparison of different methods for adapting the foundation model. Lower values are better for all metrics.}
\centering
\small
\begin{tabular}{c|ccc|c}
\hline
\textbf{Adapting Methods} & $\rm VOI_{s}$ & $\rm VOI_{m}$ &  $\rm VOI$ & Arand  \\ \hline

Frozen     & 1.318 &   0.332    &1.649  &   0.147      
\\
Fine-tuning         &1.176 &0.179   & 1.355  & 0.108     \\
Refinement          &  0.679  &  0.372  &  1.051   & 0.142    \\
\hline
Our Method & 0.614      &  0.183     & \textbf{0.797}    & \textbf{0.069}  \\
\hline
\end{tabular}
\label{tab:way_ablation}
\end{table}

\subsection{Feature-Guided Attention Analysis}
\label{sec:fga analysis}
\begin{figure*}[htbp]
    \centering
    \includegraphics[width=0.8\textwidth]{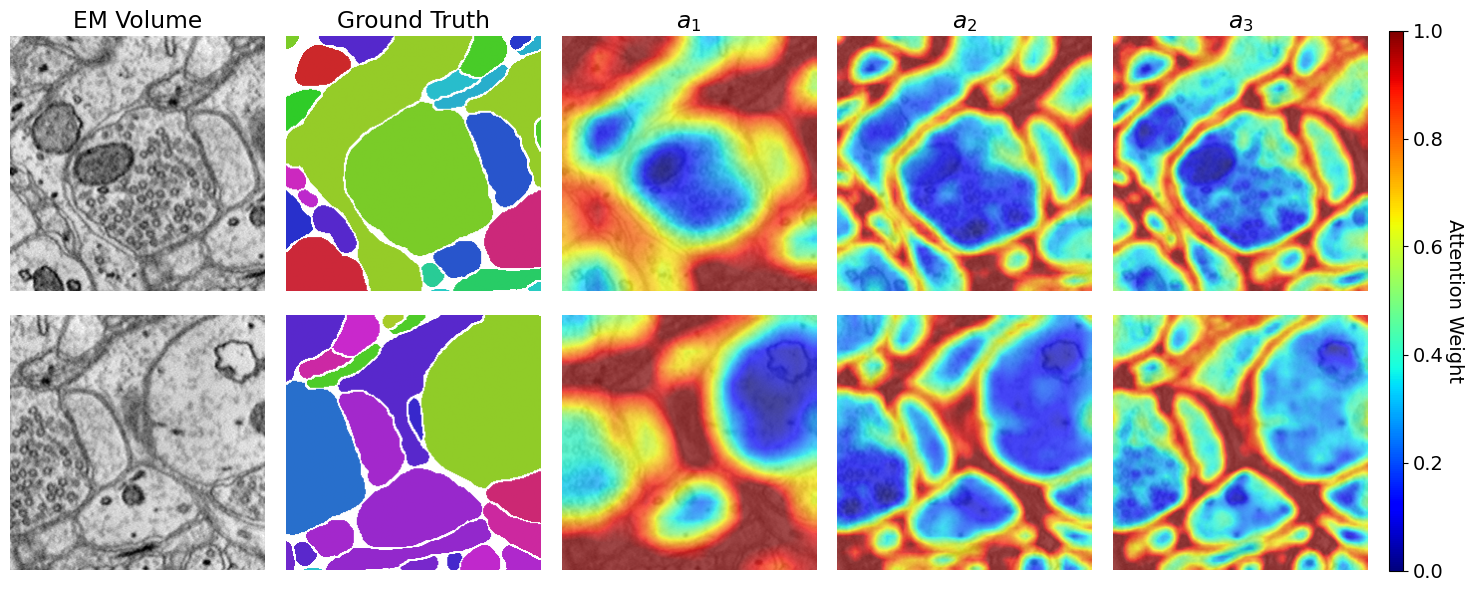}
    \caption{Hierarchical attention visualization for 3D EM neuron segmentation. Rows show sequential 2D slices. Columns 1–2 display input EM volumes and ground truth segmentations.Columns 3–5 display attention maps ($a_1$, $a_2$, $a_3$) overlaid on EM data, with warmer jet colormap colors indicating higher attention weights. A vertical color bar quantifies attention intensity, highlighting critical structural details like neuron boundaries.}
    \label{fig:attention_weights}
\end{figure*}

To address SAM2's coarse feature extraction in EM segmentation, we designed FGA module to recover the overlooked fine details (see Figure~\ref{fig:network}(b)). 
To validate our feature-guided attention mechanism, we visualize attention maps (which refer to \(a_i\) in Equation \ref{eq:a}) (Figure \ref{fig:attention_weights}) from EM neuron segmentation. The visualization presents input EM slices, ground truth segmentations and hierarchical attention maps. A jet colormap color bar quantifies attention weights, where warm colors (red/yellow) indicate higher attention and cool colors (blue/green) represent lower weights.

Key observations confirm the mechanism’s effectiveness: the attention maps exhibit a progressive refinement across \(a_1\) to \(a_3\), shifting focus from broader neurite regions to increasingly fine-grained structural details in alignment with neural structural hierarchy. Domain-specific focus is evident as high-attention regions consistently coincide with critical segmentation structures such as neuron boundaries, small neurites, and synapses, while lower attention weights are concentrated in homogeneous cell interiors and irrelevant organelles.

This discriminative attention pattern reflects EM-specific knowledge that the natural image-pretrained SAM2 encoder lacks. Our feature-guided attention mechanism enables FGE to capture these critical details, effectively addressing limitations in cross-domain transfer.

\subsection{Ablation Study} 

\begin{table}[t]
\caption{Ablation study on different modules. Lower values are better for all metrics.}
\centering
\small
\begin{tabular}{l|ccc|c}
\hline
\textbf{Model Variants} & $\rm VOI_{s}$ & $\rm VOI_{m}$ &  $\rm VOI$ & Arand  \\ \hline

w/o SAM2 Encoder     &0.682  & 0.426      &1.108  & 0.164          \\
w/o FGE         &1.176&0.179   & 1.355  & 0.108          \\
w/o FGA Block          &  0.679     &   0.372    & 1.051    &    0.142      \\ 
\hline
All Modules & 0.614      &  0.183     & \textbf{0.797}    & \textbf{0.069} \\
\hline
\end{tabular}
\label{tab:module_ablation}
\end{table}

\begin{table}[t]
\caption{Ablation study on different FGA module. Lower values are better for all metrics.}
\centering
\small
\begin{tabular}{l|ccc|c}
\hline
\textbf{FGA Type} & $\rm VOI_{s}$ & $\rm VOI_{m}$ &  $\rm VOI$ & Arand  \\ \hline

SE-Type     & 0.671 &  0.308     & 0.979 &  0.110         \\
CABM-Type        &0.608 &  0.276  &  0.883 &   0.104       \\
ECA-Type         & 0.604      & 0.235      & 0.839   &  0.091      \\
\hline
Our Method & 0.614      &  0.183     & \textbf{0.797}    & \textbf{0.069} \\
\hline

\end{tabular}
\label{tab:fga_ablation}
\end{table}

\begin{table}[t]
\caption{Ablation study on different depth of FGE. Lower values are better for all metrics.}
\centering
\small
\begin{tabular}{c|ccc|c}
\hline
\textbf{Depth} & $\rm VOI_{s}$ & $\rm VOI_{m}$ &  $\rm VOI$ & Arand  \\ \hline

3     &  0.624 & 0.192      & 0.816 & 0.090          \\
4 & 0.614      &  0.183     & \textbf{0.797}    & \textbf{0.069} \\ 
5          &  0.608     &  0.276     &  0.883   &  0.104     \\    
\hline
\end{tabular}
\label{tab:depth_ablation}
\end{table}

We conducted a series of ablation experiments on the AC3/AC4 datasets to verify the effectiveness of our proposed method.

\subsubsection{Ablation of Network Modules}
In Table~\ref{tab:module_ablation}, we further ablated different modules in the proposed network to evaluate their individual contributions. Removing any component, whether SAM2, the FGE, or the FGA, led to performance degradation, confirming the effectiveness of our proposed framework as an integrated paradigm. Specifically, the absence of FGE resulted in diminished fine-grained segmentation performance, as its refining capability is critical for capturing intricate structural details. Without the FGA, the guidance mechanism between SAM2 and FGE was disrupted, leading to suboptimal alignment of attention toward key regions. Meanwhile, removing SAM2 caused the loss of coarse feature guidance, which serves as the foundational structure for accurate segmentation. Together, these results underscore that each module plays an indispensable role in achieving the overall performance, validating the rationality of our integrated design.
\subsubsection{Comparison of FGA Implementations}
In designing our FGA module, we drew inspiration from attention mechanisms proposed in prior works. Specifically, the SE-Type~\cite{hu2018squeeze} Module achieves channel-wise attention by squeezing feature maps into global descriptors via average pooling and recalibrating channel weights through two fully connected layers. The CBAM-Type~\cite{woo2018cbam} Module extends this by incorporating both channel attention and spatial attention, which is computed via a convolution layer applied to concatenated average and maximum pooled features. The ECA-Type~\cite{wang2020eca} Module simplifies channel attention by replacing fully connected layers with a 1D convolution, adapting its kernel size to the number of channels for more efficient recalibration. 
To validate the effectiveness of our specific design, we conducted comparative experiments with these existing attention mechanisms. Experimental results in Table~\ref{tab:fga_ablation} demonstrate that our designed attention mechanism outperforms alternatives, as it is uniquely tailored to our paradigm of guiding fine-grained feature extraction using coarse features. This superiority confirms the rationality of our FGA design, which is specifically optimized for the cross-scale guidance scenario in our framework.
\subsubsection{Impact of FGE Depth}
Finally, we analyzed the impact of FGE depth in Table~\ref{tab:depth_ablation}. As a lightweight refinement network, FGE does not require excessive parameters, and a shallow architecture with only a few layers suffices. Experimental results validate that increasing network depth has minimal effect on performance, confirming that deeper layers are unnecessary for effective refinement and further supporting the efficiency of our lightweight design. This is because foundation models like SAM2 already encode a wealth of generic and coarse-grained features, allowing the refinement network to focus exclusively on capturing specific fine-grained details. Consequently, FGE can achieve excellent performance with relatively few parameters.

\section{Conclusion}

We propose FGNet, a 3D EM neuron segmentation framework that transfers knowledge from SAM2 to the EM domain. By leveraging the powerful feature extraction capabilities of SAM2 via a Feature-Guided Attention module, our method effectively bridges the domain gap between natural images and 3D EM data. Experiments validated its competitive performance with frozen SAM2 weights and significant SOTA outperformance after fine-tuning.

\section*{Acknowledgements}
This work was supported by the National Key Research and Development Program of China under
Grant 2021ZD0200301, the National Natural Science Foundation of China under Grants 62576187
and U2341228.

\bibliography{aaai2026}

\end{document}